# Collaborating Robotics Using Nature-Inspired Meta-Heuristics

M. A. El-Dosuky[1], M. Z. Rashad[1], T. T. Hamza[1], and A.H. EL-Bassiouny[2]

[1] Department of Computer Sciences, Faculty of Computers and Information sciences, Mansoura University, Egypt
mouh_sal_010@mans.edu.eg
magdi_12003@yahoo.com
Taher_Hamza@yahoo.com

[2] Department of Mathematics, Faculty of Sciences, Mansoura University, Egypt
el_bassiouny@mans.edu.eg

*Abstrac*t

This paper introduces collaborating robots which provide the possibility of enhanced task performance, high reliability and decreased. Collaborating-bots are a collection of mobile robots able to self-assemble and to self-organize in order to solve problems that cannot be solved by a single robot. These robots combine the power of swarm intelligence with the flexibility of self-reconfiguration as aggregate Collaborating-bots can dynamically change their structure to match environmental variations. Collaborating robots are more than just networks of independent agents, they are potentially reconfigurable networks of communicating agents capable of coordinated sensing and interaction with the environment. Robots are going to be an important part of the future. Collaborating robots are limited in individual capability, but robots deployed in large numbers can represent a strong force similar to a colony of ants or swarm of bees. We present a mechanism for collaborating robots based on swarm intelligence such as Ant colony optimization and Particle swarm Optimization

**Keywords: Collaborating Robotic, swarm intelligence, Artificial Life, Ant colony optimization, Particle swarm Optimization, environment**

**I. INTRODUCTION**

Collaborating robotics is currently one of the most important application areas for swarm intelligence. Swarms provide the possibility of enhanced task performance, high reliability (fault tolerance), low unit complexity and decreased cost over traditional robotic systems. They can accomplish some tasks that would be impossible for a single robot to achieve. Collaborating-bots are a collection of mobile robots able to self assemble and to self organise in order to solve problems that cannot be solved by a single robot. These robots combine the power of swarm intelligence with the flexibility of self reconfiguration as aggregate Collaborating-bots can dynamically change their structure to match environmental variations[1]. Collaborating robots are more than just networks of independent agents they are potentially reconfigurable networks of communicating agents capable of coordinated sensing and interaction with the environment.

**Insects Software**

Collaborating robots main idea is based on Local interactions between nearby robots, which are being used to produce large scale group behaviors from the entire swarm. Ants , bees and termites are beautifully engineered examples of this kind of software in use. These insects do not use centralized communication; there is no strict hierarchy, and no one in charge. However, developing swarm software from the "top down", i.e., by starting with the group application and trying to determine the individual behaviors that it arises from, is very difficult. Instead a "group behavior building blocks" that can be combined to form larger, more complex applications are being developed.[2]

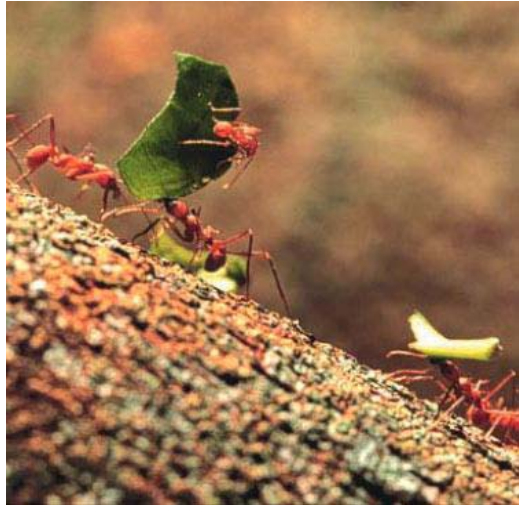

**Figure 1: Leaf Cutter Ants[8]**

**Robot types:**
There are many applications for collection of robots. In all these applications, collaborating robots must work independently, only communicating with other nearby robots. It is either too expensive (robot vacuums need to be very cheap, too far (it takes 15 minutes for messages to get to Mars), or impossible (radio control signals cannot penetrate into earthquake rubble) to control all of the robots from a centralized location. However, a distributed control system can let robots from a centralized location. However, a distributed control system can let robots interact with other nearby robots, cooperating amongst themselves to accomplish their mission.

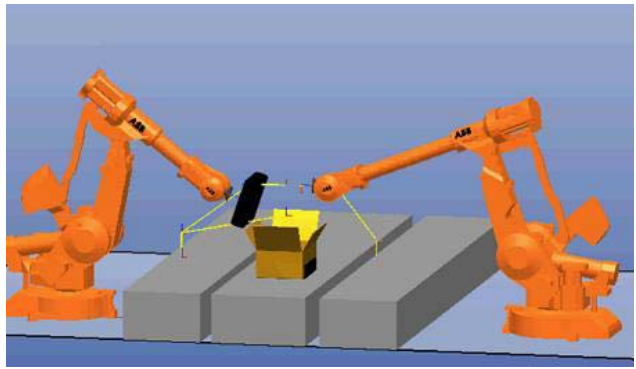

**Figure2: Co-operative Robots as automatic assembly [9]**

Collaborating robots come in many forms. **Modular robots** are made of lots of small, identical modules. A module is essentially a small, relatively simple robot or piece of a robot. **Chain robots** are long chains that can connect to one another at specific points. The basic idea of a **lattice robot** is that robots of small, identical modules that can combine to form a larger robot.([7], [10])

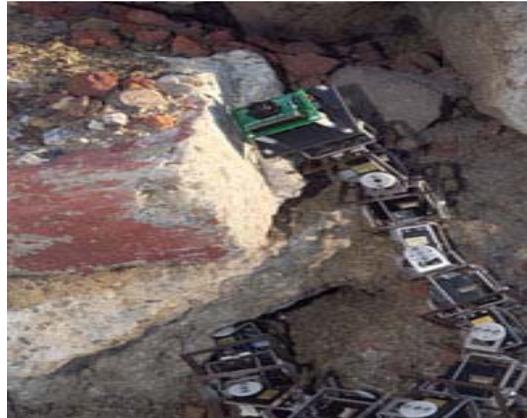

**Figure 3. NASA's example of a chain robot. [10]**

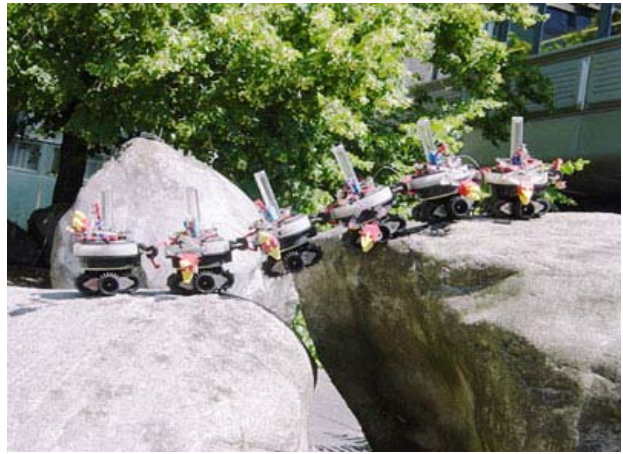

**Figure 4 Collaborating-bots can maneuver independently, or they can combine to complete tasks. [10]**

## II. METHODOLOGY

The general architecture of the control unit of most robots is shown in the next figure.

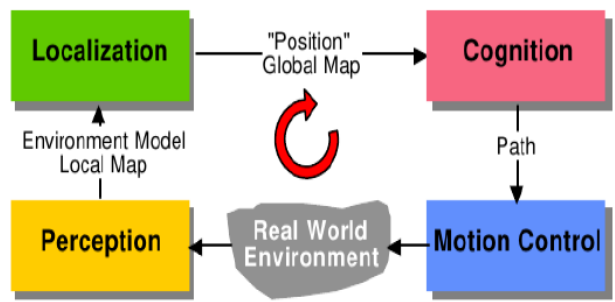

Figure 4 Control Architectures ([5])

    Collaborating robots can locate and form a perimeter around an area in the task environment. The length of time necessary for the first robot to locate and identify the area differed significantly, since the robots initially perform a **random search**. Once one robot found the proper area and began to emit an audible signal, congregation and perimeter formation occurred quickly as the other robots were drawn to the "come hither" chirp.

For environment representation, there are three forms: Continuous Metric, Discrete Metric (also known as Metric grid) and Discrete Topological (or Topological grid). We recommend the Discrete Topological because it suits the collaborating robots, since the collaborating robots form a topology themselves.[5]

Next we give a summary of the methodology needed for collaborating robots. Main techniques are Ant colony optimization and Particle swarm Optimization

**Ant colony optimization:**
Ant colony optimization or ACO is a Meta heuristic optimization algorithm that can be used to find approximate solutions to difficult combinatorial optimization problems. In ACO artificial ants build solutions by moving on the problem graph and they, mimicking real ants, deposit artificial pheromone on the graph in such a way that future artificial ants can build better solutions. ACO has been successfully applied to an impressive number of optimization problems[3].

**Particle swarm Optimization:**
Particle swarm optimization or PSO is a global optimization algorithm for dealing with problems in which a best solution can be represented as a point or surface in an n-dimensional space. Hypotheses are plotted in this space and seeded with an initial velocity, as well as a communication channel between the particles. Particles then move through the solution space, and are evaluated according to some fitness criterion after each time step. Over time, particles are accelerated towards those particles within their communication grouping which have better fitness values. The main advantage of such an approach over other global minimization strategies such as simulated annealing is that the large numbers of members that make up the particle swarm make the technique impressively resilient to the problem of local minima.[4]

**Design recommendations:**
Now let us talk about the best design we recommend for the collaborating robot. Higher-level layers can subsume the roles of lower levels by suppressing their outputs. However, lower levels continue to function as higher levels are added. The result is a robust and flexible robot control system. For the system to be used to control a mobile robot wandering around unconstrained laboratory areas and computer machine rooms, we recommend the TETwalker, which stands for tetrahedral walker. It is a design for a robot that uses a unique form of locomotion. A bug like robot inspired by insects that skate across water has been engineered. The machine provides deeper insight into these long legged bugs known as water striders or pond skaters move. Three flexible joint-like connections called actuators, one on the body and one where each side leg give the robot the flexibility it needs to move.[6]

A group of robots can perform navigating over an area of unknown terrain over a target light source. If possible, the robots should navigate to the target independently. If, however, the terrain proves too difficult for a single robot, the group should self-assemble into a larger entity and collectively navigate to the target.

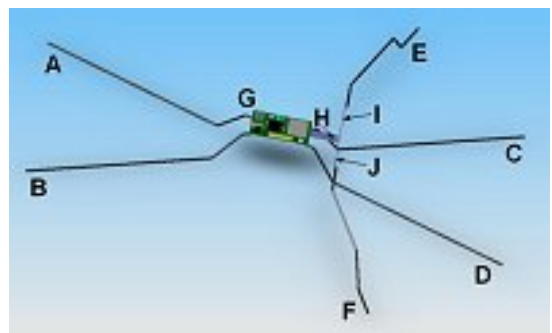

**Figure 5 mosquito-like TETwalker**
A, B, C and D are the supporting legs; E and F are the actuating legs; G is the body with sensors, power sources and a wireless communication module; H is the middle actuator; and I and J are the right/left actuators [6]

**Algorithm of Pheromone Control**
Pheromone control adopts several approaches to reduce the influences from past experience and encourages the exploration of new paths or paths that were previously non-optimal. The approaches as follows:
1) Evaporation
2) Aging

3) Limiting and Smoothing Pheromone
4) Pheromone-Heuristic Control

**1. Evaporation**
To reduce the effect of past experience, an approach called evaporation is typically used in conjunction with ACO. Evaporation prevents pheromone concentration in optimal paths from being excessively high and preventing ants from exploring other (new or better) alternatives. In each iteration, the pheromone values $T_{ij}$ in all edges are discounted by a factor such that

$$T_{ij} \leftarrow T_{ij}(1-p).$$

Suppose that at some iteration $t_i$, all ants converge to a path $R_i$, and deposit a very high concentration of pheromone (represented with larger triangles). In the next iteration $t_{i+1}$, the pheromone concentration along $R_i$ is reduced by some factor (represented by smaller triangle), and at $t_{i+2}$, the pheromone concentration is further reduced.

**Implementation of Evaporation**
```
'Update Pheromone Value
For i = 0 To ptop - 1
pstrength(i) = pstrength(i) * 0.9
Next
```

**2 Aging**
Additionally, past experience can also be reduced by controlling the amount of pheromone deposited for each ant according to its age. This approach is known as *aging*. In aging, an ant deposits lesser and lesser pheromone as it moves from node to node.
Aging is based on the rationale that "old" ants are less successful in locating optimal paths since they may have taken longer time to reach their destinations. Both aging and evaporation include recency as a factor of routing preference, hence, if a favorable path is not chosen recently, its preference will be gradually eliminated. By making existing pheromone trail less significant than the recent pheromone updates, both aging and evaporation encourage discoveries of new paths that were previously nonoptimal.

**Implementation of Aging**
```
agefactor = 0.9
updted = 0
For I = 0 To ptop - 1
If ppath(I) = path Then
   updted = 1
   pstrength(I) = pstrength(I) + agefactor ^ hops
End If
Next
If updted = 0 Then
   ppath(ptop) = path
   pstrength(ptop) = agefactor ^ hops
   ptop = ptop + 1
End If
```

**Limiting and Smoothing Pheromone**
Stagnation by *limiting* the amount of pheromone in every path. By placing an upper bound $T_{max}$ on the amount of pheromone for every edge $(i, j)$, the preference of an ant for optimal paths over nonoptimal paths is reduced. This approach prevents the situation of generating a dominant path. A variant of such an approach is *pheromone smoothing*. Using pheromone smoothing, the amount of pheromone along an edge is reinforced as follows:

$$T_{ij}(new) = T_{ij}(old) + \delta * (T_{max} - T_{ij}(old))$$

where δ is a constant between 0 and 1.
It can be seen that as $T_{ij}(old) \to T_{max}$ a smaller amount of pheromone is reinforced along an edge $(i, j)$. Although not totally identical, pheromone smoothing also bears some resemblance to evaporation. While evaporation adopts a uniform discount rate for every path, pheromone smoothing places a relatively greater reduction in the reinforcement of pheromone concentration on the optimal path(s). Consequently, pheromone smoothing seems to be more effective in preventing the generation of dominant paths.

**Implementation of Limiting and Smoothing Pheromone**
```
    tmax = 100
    smoothing = 0.33
    Private Sub updperval(path As String)
    updted = 0
    For i = 0 To ptop - 1
    If ppath(i) = path Then
        updted = 1
        pstrength(i) = pstrength(i) + smoothing * (tmax - pstrength(i))
    End If
    Next
    If updted = 0 Then
        ppath(ptop) = path
        pstrength(ptop) = smoothing * (tmax)
    List4.AddItem pstrength(ptop)
        ptop = ptop + 1
    End If
```

**4 Pheromone-Heuristic Control**

Another approach is to configure ants so that they do not solely rely on sensing pheromone for their routing preferences. This can be accomplished by configuring the probability function $P_{ij}$ for an ant to choose an edge (i , j) using a combination of both pheromone concentration $T_{ij}$ and heuristic function $\eta_{ij}$. As noted , an ant selects an edge probabilistically using $T_{ij}$ and $\eta_{ij}$ as a functional composition for $P_{ij}$. In network routing, $\eta_{ij}$ is a function of the cost of edge (i , j)(which may include factors such as queue length, distance, and delay).

$\alpha$ and $\beta$ represent the respective adjustable weights of $T_{ij}$ and $\eta_{ij}$. Consequently, the routing preferences of ants can be altered by selecting different values of $\alpha$ and $\beta$. If $\alpha > \beta$ ants favor paths with higher pheromone concentrations, and a higher value of $\beta$ directs ants to paths with more optimistic heuristic values. In general, different values of $\alpha$ and $\beta$ are suitable to be applied at different states of a network. A lower value of $\alpha$ is generally preferred when pheromone concentration along paths may not necessarily reflect their optimality. Examples of such situations include the initial stage after a network reboots (before the network stabilizes), and when there are frequent and abrupt changes in network status due to either link (or node) failure or introduction of new paths (nodes). However, as a network stabilizes, a higher value of $\alpha$ is preferred. Furthermore, recent research demonstrated that dynamically altering the values of $\alpha$ and $\beta$ in response to changes in network status may increase the performance of ants.

**Implementation of Pheromone-Heuristic Control**
```
    For i = 0 To List2.ListCount - 1
        If Left$(List2.List (i), 13) = stringparse (b, "DEST=") Then
        commaparse1 (List2.List (i))
    reddo1:
        l = Int (Rnd * iptop1 + 1)
        If l = 0 Then GoTo reddo1
        tosys = ips1 (l)
        End If
    Next
    i = Int (Rnd * iptop1)
    i = i + 1
    pher:
    small = Len (ppath(0))
    locn = 0
    For i = 0 To ptop - 1
    If Len (ppath (i)) < small Then
        small = Len (ppath(i))
        locn = i
    End If
    Next
    pstrength (locn) = pstrength (locn) + 1
```

## III. RESULTS

By running the algorithm, we get the following results
Between the magnitude of average velocity (|v|) and the ordering factor (Δ). Where

$$V = \sum v_i / N$$

And $v_i$ is the velocity of ant number i and N is the total number of ants. The ordering factor (Δ) can be measured by dividing the number of ants matching the desired topology over the total number of ants.

Table 1: relation between ordering factor (Δ) and magnitude of average velocity (|v|)

| ordering factor (Δ) | magnitude of average velocity (|v|) |
|---|---|
| 0.05 | 0.9 |
| 0.10 | 0.86 |
| 0.15 | 0.81 |
| 0.20 | 0.73 |
| 0.30 | 0.62 |
| 0.35 | 0.41 |
| 0.40 | 0.001 |
| > 0.40 | 0 |

## IV. DISCUSSION

Figure 6 shows the relation between ordering factor (Δ) and magnitude of average velocity (|v|).

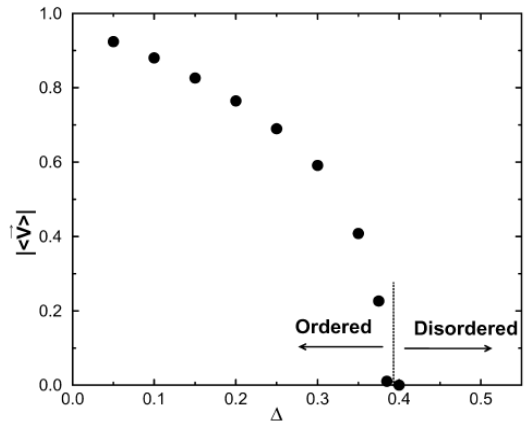

Figure 6: relation between ordering factor (Δ) and magnitude of average velocity (|v|)

We recommend avoiding increasing the desired ordering factor over 40% of the desired topology, as the collaborating robots will suddenly stop.

## V. CONCLUSION

This paper introduces collaborating robots which provide the possibility of enhanced task performance, high reliability and decreased. Collaborating-bots are a collection of mobile robots able to self-assemble and to self-organize in order to solve problems that cannot be solved by a single robot. These robots combine the power of swarm intelligence with the flexibility of self-reconfiguration as aggregate Collaborating-bots can dynamically change their structure to match environmental variations.

We present a mechanism for collaborating robots based on swarm intelligence such as Ant colony optimization and Particle swarm Optimization. Robots are going to be an important part of the future. Once robots are useful, groups of robots are the next step, and will have tremendous potential to benefit mankind. Software designed to run on large groups of robots is the key needed to unlock this potential.